# Data-Efficient Machine Learning Potentials *via* Difference Vectors Based on Local Atomic Environments

Xuqiang Shao[a, b, †], Yuqi Zhang[a, b, †], Di Zhang[c, †], Zhaoyan Dong[a, b], Tianxiang Gao[d], Mingzhe Li[d] Xinyuan Liu[d], Zhiran Gan[d], Fanshun Meng[e], Lingcai Kong[f], Zhengyang Gao[d], Hao Li[c*], Weijie Yang[d*]

[a] Department of Computer, School of Control and Computer Engineering, North China Electric Power University, Baoding 071003, China

[b] Engineering Research Center of Intelligent Computing for Complex Energy Systems, Ministry of Education, Baoding, Hebei 071003, China

[c] Advanced Institute for Materials Research (WPI-AIMR), Tohoku University, Sendai 980-8577, Japan

[d] Department of Power Engineering, School of Energy Power and Mechanical Engineering, North China Electric Power University, Baoding, 071003, Hebei, China

[e] School of Science, Liaoning University of Technology, Jinzhou, 12100l, China

[f] Department of Mathematics and Physics, School of Mathematics and Physics, North China Electric Power University, Baoding, 071003, Hebei, China

† These authors contributed equally to this work.

* Correspondence and requests for materials should be addressed to:

W. Y. (yangwj@ncepu.edu.cn)

H. L. (li.hao.b8@tohoku.ac.jp).

**Abstract**

Constructing efficient and diverse datasets is essential for the development of accurate machine learning potentials (MLPs) in atomistic simulations. However, existing approaches often suffer from data redundancy and high computational costs. Herein, we propose a new method—Difference Vectors based on Local Atomic Environments (DV-LAE)—that encodes structural differences *via* histogram-based descriptors and enables visual analysis through t-SNE dimensionality reduction. This approach facilitates redundancy detection and dataset optimization while preserving structural diversity. We demonstrate that DV-LAE significantly reduces dataset size and training time across various materials systems, including high-pressure hydrogen, iron–hydrogen binaries, magnesium hydrides, and carbon allotropes, with minimal compromise in prediction accuracy. For instance, in the α-Fe/H system, maintaining a highly similar MLP accuracy, the dataset size was reduced by 56%, and the training time per iteration dropped by over 50%. Moreover, we show how visualizing the DV-LAE representation aids in identifying out-of-distribution data by examining the spatial distribution of high-error prediction points, providing a robust reliability metric for new structures during simulations. Our results highlight the utility of local environment visualization not only as an interpretability tool but also as a practical means for accelerating MLP development and ensuring data efficiency in large-scale atomistic modeling.

**Introduction**

In the cutting-edge research of modern materials science and chemical physics, first-principles calculations, with their theoretical rigor, provide a solid basis for revealing the physical and chemical properties at the atomic scale, and becomes an indispensable tool for understanding and designing new materials.[1–3] However, these methods based on quantum mechanics, such as density functional theory (DFT), while accurate, are limited by high computational costs, especially for complex systems containing a large number of atoms and long-term dynamics simulations, which significantly limit their application.[4,5] It is in this context that machine learning potentials (MLPs) came into being: by learning the reference data obtained from first-principles calculations, they can reconstruct the interatomic potential energy surface with high efficiency and near-first-principles accuracy, dramatically widening the space-time scale of the simulation, and thus showing extraordinary potential in understanding chemical reaction mechanisms, phase transition processes, and material property predictions.[6–11]

The construction of MLPs is an intricate endeavor, predominantly contingent upon **two pivotal components**: the design of the ML model and the construction of the dataset. Collectively, these elements dictate the ultimate performance, precision, and generalizability of the model.[12–14] An adeptly crafted ML model is essential for the efficacious capture and articulation of the underlying physical laws. In the construction of deep neural network models, a series of critical requirements must be met, which are not limited to the model having an appropriate functional form, maintaining symmetry, and possessing good scalability.[15,16] The meticulous design of the model architecture, including number of layers, number of nodes, and selection of activation functions, is crucial for determining the model's complexity, training efficiency, and its ability to fit the complex potential energy surface (PES).[17,18] These characteristics collectively ensure that the model can accurately predict the interactions and potential energies under different atomic structures.

Despite the plethora of research dedicated to the refinement of ML models, culminating in the emergence of various high-performance network models, such as MACE[19] focusing on higher-order equivariant interactions and Allegro[20] which implements efficient E(3) equivariant graph neural networks for large-scale atomistic dynamics, the dataset construction still remains a crucial component. In the absence of a high-quality dataset, even the most sophisticated models are impeded from realizing their full potential. The construction of a dataset is the linchpin in ensuring the model's generalizability. In the context of MLP modeling, this typically necessitates the assembly or simulation of a vast array of atomic configurations that encapsulate the myriad states the system may inhabit, alongside the acquisition of their corresponding energies and forces. An ideal dataset ought to span the entire phase space of the system, guaranteeing that the model learns most of the pivotal physical characteristics during the training process. Should the dataset be deficient or biased, the model may succumb to overfitting to specific patterns within the data, thereby significantly undermining its predictive accuracy in previously unseen scenarios.[21–23]

In contemporary materials science research, molecular dynamics (MD) and Monte Carlo (MC) simulations have become important means of generating data sets of structures, energies, and forces.[24,25] These simulation techniques provide rich data resources for the training of ML models, especially in the era of big data, where publicly available material databases greatly expand the scale and diversity of available data.[26,27] However, although the surge in data volume provides a broader space for model training, it also raises many challenges. In practice, it is not practical to use all data for training, so it is crucial to accurately select the structural data that meets the research objectives.

Although the traditional manual screening method can be integrated into the in-depth understanding of experts, it has some inherent limitations.[28] First, the manual screening process is not only time-consuming but also computationally expensive, especially when dealing with large data sets. Secondly, the method is susceptible to subjective judgment, which may lead to selection bias and inconsistency of results. More critically, manual filtering may not be able to fully cover the entire data set for making full use of data resources. Additionally, the use of principal component analysis (PCA) for dimensionality reduction,[29] a common approach in data preprocessing, has its own set of disadvantages. PCA, while effective in capturing global patterns, often overlooks local structural nuances that are critical for material science applications. **It tends to smooth over the subtle variations that could be pivotal for understanding complex material behaviors, leading to a loss of critical information and potentially misleading interpretations.** To address these limitations, nonlinear dimensionality reduction techniques such as t-SNE (t-Distributed Stochastic Neighbor Embedding) have gained prominence.[30,31] Unlike PCA, t-SNE prioritizes the preservation of local structural similarities in high-dimensional data through probabilistic neighborhood matching, making it particularly effective for visualizing clusters or identifying fine-grained patterns in complex material systems. However, its computational scalability and sensitivity to parameter selection (*e.g.*, perplexity) remain challenges in large-scale applications.

To break through the bottleneck of conventional data screening, researchers gradually adopt active learning strategies such as the DP-GEN.[32] The DP-GEN framework combines the advantages of active learning and MD simulations to intelligently guide the selection of data samples through the real-time evaluation of model uncertainty. This strategy not only significantly reduces the reliance on manual labeling but also improves the efficiency and accuracy of the screening process. DP-GEN iteratively selects high-uncertainty configurations from MD simulations and augments the training set with labeled quantum mechanics data, thereby refining the potential energy model for reliable material simulations. However, the computational cost of the DP-GEN iterative process (due to repeated quantum mechanics calculations) is relatively high. Therefore, how to effectively mine and apply high-quality data in large datasets at low cost remains an important and urgent challenge in the field.

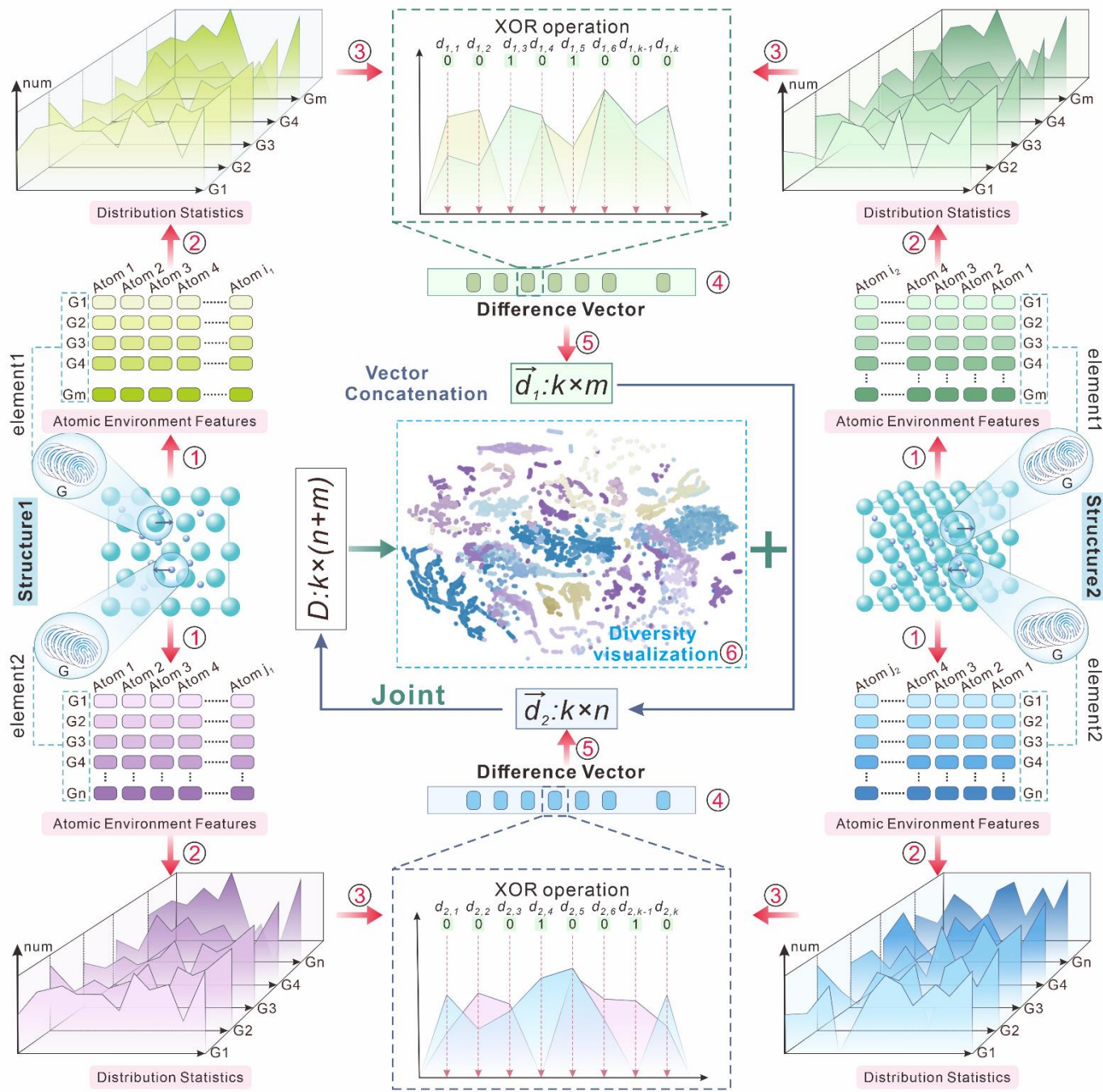

**Figure 1. Schematic workflow for constructing structural difference vectors and optimizing datasets in a binary system.** ① Calculate Atomic Environment Features: for each atom in the

structure, compute descriptors (*e.g.*, symmetry functions, G) based on its local neighborhood. ② Generate Distribution Statistics: for each descriptor type, create a histogram (with *k* bins) summarizing its value distribution across all atoms in the structure. ③ Perform Exclusive OR (XOR) Operation: compare the histogram bins of the current structure with a reference structure using XOR (as defined in Eq. 5 of the **Methods section**) to identify differing intervals ('1' for difference; '0' for similarity). ④ Generate Difference Vector (Descriptor Specific): aggregate the XOR results (0s and 1s) across all *k* bins for a single descriptor into a binary difference vector (of length *k*) specific to that descriptor. ⑤ Vector Concatenation: concatenate all the descriptor-specific difference vectors (from Step ④). For a binary system as depicted, if element 1 has m descriptor types and element 2 has n descriptor types, this step forms respective matrices $\vec{d_2}$ (size $k \times m$) and $\vec{d_2}$ (size $k \times n$). These are then concatenated to form a single, comprehensive joint difference matrix D (size $k \times (m + n)$) representing the overall structural difference from the reference. ⑥ Diversity Visualization: use dimensionality reduction (*e.g.*, t-SNE) on the comprehensive difference vectors of multiple structures to visualize their similarity and diversity in a low-dimensional space.

Motivated by the current stages and challenges described above, in this study, we develop an innovative method to construct structural difference vectors by integrating features and descriptors of the local atomic environment using histogram statistics, as outlined in **Figure 1**. These difference vectors are then leveraged for data-efficient dataset screening and optimization. Their relationship can also be reduced and visualized *via* the t-SNE algorithm to provide insights into the dataset's diversity and structure. **This novel method significantly reduces dataset redundancy while maintaining data diversity, effectively improving the training efficiency of MLPs without sacrificing the model's generalization ability.**

The core idea behind the DV-LAE method, illustrated in **Figure 1**, is to find a better way to compare atomic structures quickly. Traditionally, this often involves checking every detail of high-dimensional atomic environment descriptors, which is computationally intensive and potentially sensitive to minor structural variations. **Instead, DV-LAE employs a statistical representation of these environments.** By first converting atomic environment features (①) into a statistical distribution, like a bar chart or histogram (②), it summarizes the overall "flavor" or main characteristics of the local atomic neighborhood. Then, to compare the summaries of two structures, a key step uses a clever binary comparison (③), acting like a simple "Yes/No" question. Rather than measuring the exact numerical difference between the charts, this simply checks if there is a statistically significant difference, not just random variation ('1' for a meaningful difference, '0' for similarity, as detailed in the **Methods section**). This process, which turns comparisons into simple '1's and '0's, creates a compact and fast representation of structural dissimilarity, focusing on where differences exist for quick initial checking.

These '1's and '0' results from comparing different feature types are collected into short "difference lists" (④), which are then combined into one comprehensive difference vector (⑤). This final vector acts like a simple code showing in what specific ways a structure meaningfully differs from a reference structure. This difference representation is designed to highlight structural changes, making it very effective for quickly finding entirely new or rare shapes within a large dataset. Furthermore, by creating visual maps (⑥) from these comprehensive difference vectors, we can see the diversity of the entire dataset at a glance. This helps identify groups of very similar structures (redundancy) or spot entirely separate families of structures. **Overall, the goal of the DV-LAE method is to make the datasets used for training MLPs more efficiently by cutting down on unnecessary duplicates while ensuring a good variety of structures is kept.** This makes training faster without hurting the model's accuracy on new structures. The subsequent sections will present the performance of this DV-LAE method across various material systems and discuss its utility in materials analyses.

**Results and Discussion**

2.1 Structural Difference Vector Analysis for Discriminative Material Sample Visualization

Visual analysis of sample distribution is of great significance in materials research. Current mainstream methods rely on dimensionality reduction of material descriptors for data visualization. A critical prerequisite for such analysis is the construction of robust structural descriptors that encode atomic configurations into mathematically tractable forms. Among these descriptors, the G-function (short for structural descriptor function) plays a central role in quantifying local atomic environments.[33] Typically defined through radial distribution functions (RDFs) or angular distribution functions (ADFs), G-functions map the spatial arrangement of neighboring atoms into high-dimensional vectors that preserve critical structural features such as bond lengths, coordination numbers, and symmetry patterns. These descriptors have become indispensable tools for tasks ranging from phase classification to defect characterizations in materials informatics.

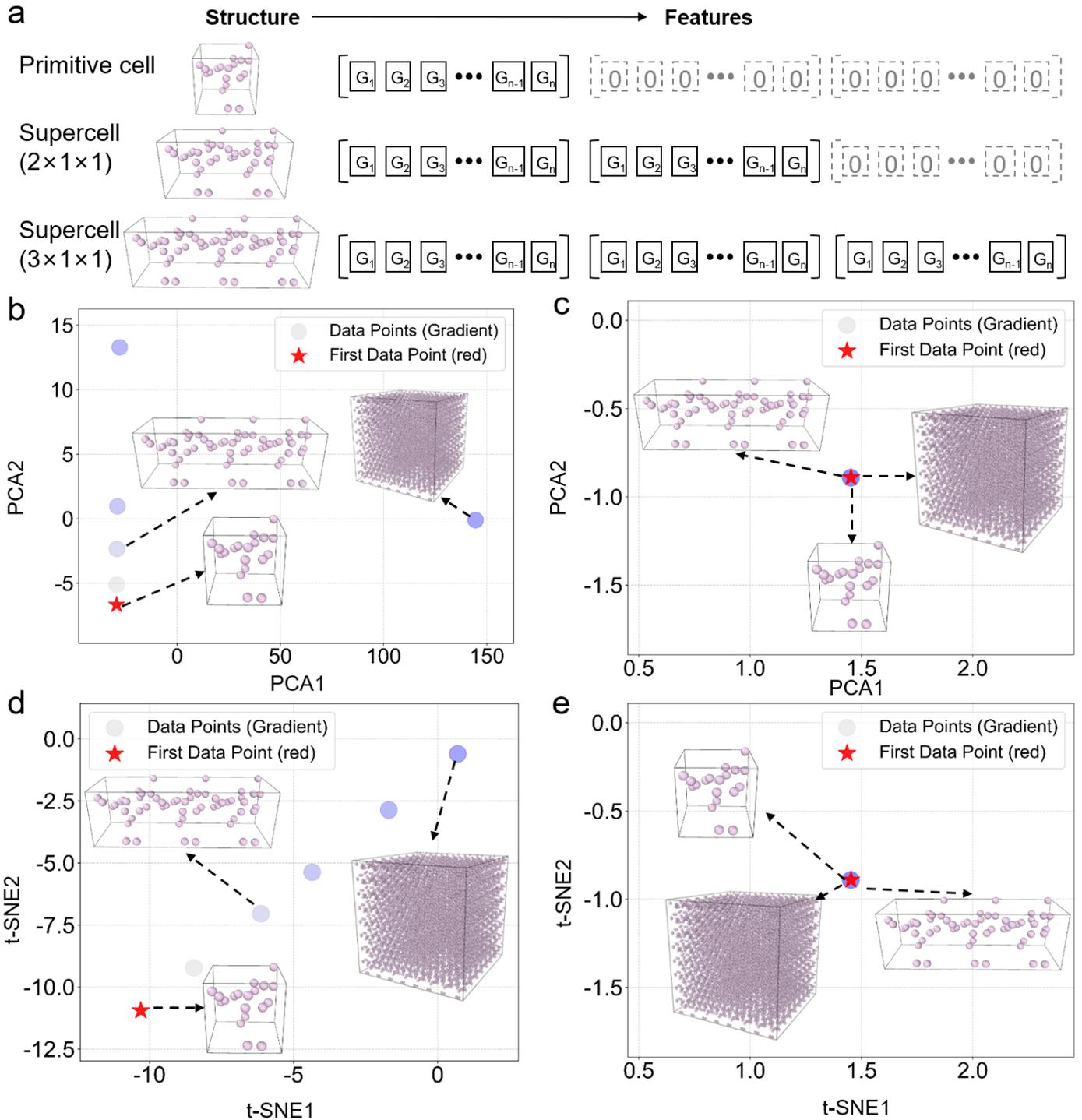

**Figure 2. The representation of supercell structures using G-function and difference-vector descriptors and comparison of their effectiveness through PCA and t-SNE visualization.** (a) Schematic illustration of the primitive cell to supercell transformation *via* G-function descriptors with dimensionless zero-padding. (b) G-function-based PCA projection visualization for supercell dataset. (c) Difference-vector-based PCA projection visualization for supercell dataset. (d) G-function-based t-SNE embedding visualization for supercell dataset. (e) Difference-vector-based t-SNE embedding visualization for supercell dataset.

However, due to the significant differences in atomic number and elemental composition of different material systems, conventional PCA is unable to directly deal with global descriptors in non-uniform dimensions. Consequently, researchers typically undertake a preprocessing step wherein they calculate the descriptor means of the same element and then integrate them with the descriptor sequences of different elements. This approach facilitates the conversion of different structural information into vectors of uniform dimensions, which are then analyzed using PCA to reduce dimensionality.

As illustrated on the left side of **Figure 2a**, G-function descriptors were constructed for three samples of hydrogen systems, where the first sample is a primitive cell and the last two are supercell structures. Due to the variation in the number of atoms, the characteristic dimensions of the G-functions of each system differ significantly. To realize the visualization requirements of the subsequent dimensionality reduction analysis, the traditional method adopts zero-padding to dimensionally align the low-dimensional G-functions (right side of **Figure 2a**). **However, this method is inherently flawed.** As illustrated in **Figure 2b**, after zero-padding treatment of six supercell samples extended by the same single cell, the 2D projections obtained by PCA demonstrate that the original single cell (red pentagrams) is distinctly separated from the supercell samples (blue circles) in the feature space. This phenomenon is contrary to the periodic nature of the material system, as theoretically, the single cell and its corresponding supercell should be characterized by identical physical properties.

To overcome the limitations of conventional methodologies, this study proposes a feature reconstruction method based on difference vector statistics. The method constructs a unified dimensional differential vector descriptor by extracting the numerical distribution of G-function distribution features. As shown in **Figure 2c**, when using PCA for dimensionality reduction, the original single cell and supercell samples represented by the new descriptors completely overlap in the feature space, accurately reflecting the consistency of their physical nature.

Beyond PCA, t-SNE is another popular technique known for its effectiveness in visualizing local structures in high-dimensional data. However, if t-SNE was applied to the traditional zero-padded G-functions, it would still inherit the descriptor's flaw, leading to an incorrect representation (as shown in **Figure 2d**), even if the visual pattern differs from PCA's result in **Figure 2b**. In contrast, **Figure 2e** demonstrates the visualization using t-SNE applied to the descriptors generated by our proposed difference vector statistics method. This also successfully captures the consistency between the primitive cell and its supercells, further validating the effectiveness of the difference vector approach in characterizing the essential nature of the material system across different visualization techniques.

The experimental results demonstrate that, in comparison with the conventional G-function directed dimensionality reduction method, the statistical difference vector method proposed in this study can more effectively characterize the essential characteristics of material systems. This robust descriptor, when paired with suitable dimensionality reduction techniques such as PCA or t-SNE, provides a reliable solution for the visualization analysis of high-dimensional features of materials.

2.2 Applications on High-Temperature, High-Pressure Hydrogen Systems

We utilized the training dataset from literature,[34] which encompasses a total of 25,440 structural configurations of the hydrogen systems under high-temperature and high-pressure conditions. This choice allows us to specifically evaluate the effectiveness of our method for systems with a single type of element, providing valuable insights into its applicability and potential limitations. By constructing difference vectors and subsequently applying t-SNE for visualization purposes, we achieved a graphical representation of the dataset's diversity (**Figure 3a**). In **Figure 3a**, the red and yellow-green regions indicate areas where a large number of structural points are close together or overlap, indicating a high density of data points. Through the analysis here, a considerable number of structural overlaps are revealed, which implies data redundancy.

Leveraging the outcomes of the visualization, we executed a targeted data screening. Consequently, we established an optimized subset comprising 17,352 structures, signifying a reduction in data volume by ~31.79%. **Figure 3b** illustrates the comparison of the average training time per iteration between the original and optimized datasets. The experiments were conducted using an AMD EPYC 7662 64-Core Processor, and this configuration was used for all subsequent computational experiments discussed herein. The average training time per iteration decreased from 29,585.8 seconds to 21,018.9 seconds, **resulting in a remarkable 29.0% reduction in time cost per iteration.** This optimization significantly diminished the expenses incurred during the training process. We utilized relevant data, screened an optimized subset, and the optimized effect is visualized in **Figure S1**, which shows the structural features of the optimized high-temperature and high-pressure hydrogen system dataset under t-SNE visualization.

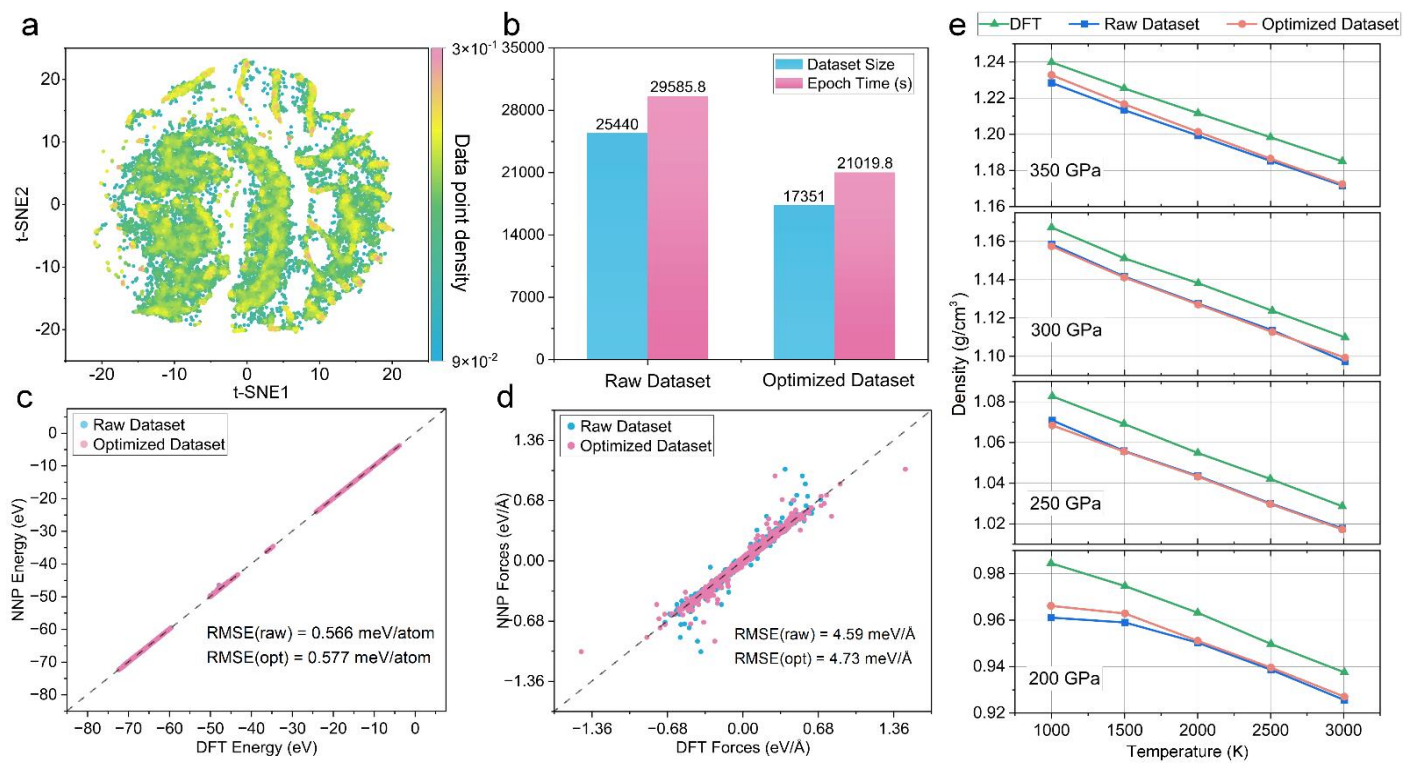

**Figure 3. Results of the dataset optimization process, comparing the original and optimized datasets in terms of data distribution, training efficiency, and prediction accuracy.** (a) t-SNE visualization of the high-temperature, high-pressure hydrogen system dataset, highlighting data point density distribution in the dataset. (b) Comparison of the average training time per iteration for the original dataset (25,440 structures) and the optimized dataset (17,352 structures). (c-d) Training results comparison: energy RMSE for the full dataset is 0.566 meV/atom and for the optimized subset is 0.577 meV/atom. Force RMSE for the full dataset is 4.59 meV/Å, and for the optimized subset is 4.73 meV/Å. (e) Comparison of density isoclines predicted by the MLP. The green triangles represent standard reference data, while the blue squares and red circles represent predictions from the MLP trained on the original dataset and the optimized dataset, respectively. Both datasets show similar accuracy.

To confirm that the optimized dataset can effectively substitute the original dataset in terms of diversity, we evaluated the prediction accuracy of energies and forces using the same test set for MLPs trained on both the original and optimized datasets. **Figures 3c-d** illustrate a comparative analysis of the predicted energies and forces using the MLP trained on these datasets. Each scatter plot aligns DFT reference values along the *x*-axis and the ML model's predictions along the *y*-axis. The red and blue spheres in these plots distinguish the predictions from the optimized and original datasets, respectively, facilitating a direct comparison of the model's performance. As shown in **Figures 3c-d**, the results reveal that the MLP trained on the optimized dataset maintains comparable accuracy to the one trained on the full dataset. In terms of energy prediction, the root mean square error (RMSE) shows a slight uptick from 0.566 to 0.577 meV/atom, indicating a mere 1.94% increase. For force prediction, the RMSE marginally increases from 4.59 to 4.73 meV/Å, representing a 3.05% increment. Despite these slight variations in RMSE, the tangible advantages of the reduced data volume and shorter training times far outweigh these minor differences, emphasizing the efficacy and utility of the new data optimization method.

To further verify the predictive ability of the MLP, we screened the data set and retrained the MLP using the optimized data. A series of comparative experiments covering the pressure ranging from 200 to 350 GPa were conducted using a method similar to the high temperature-pressure simulation described in literature.[34] The tests were performed at 50 GPa intervals, with initial temperatures from 1000 to 3000 K at each pressure point to examine temperature effects on hydrogen's equation of state. As shown in **Figure 3e**, the left panel displays density isoclines predicted by the MLP from literature.[34] The green triangles represent standard reference data, while blue squares and red circles respectively denote predictions from MLPs trained on original and optimized datasets. Both MLP versions show minimal deviation from reference data, with the optimized-dataset MLP exhibiting closer agreement at 200 GPa. This demonstrates the optimized dataset serves as an effective substitute for the original data in MLP training while maintaining the predictive accuracy.

2.3 Applications on Binary Systems (α-Fe/H)

To investigate the generalizability of our method, we further utilized the raw dataset of the α-Fe/H binary systems provided by Ref.[35], which contains 19,743 structural configurations. As demonstrated in **Figure 4a**, the use of the t-SNE method for dimensionality reduction and visualization of difference vectors reveals that structures mapped in different colors are densely clustered, indicating a high degree of similarity among numerous structures in expressing the diversity of the system, thereby highlighting significant data redundancy. Based on this observation, employing the aforementioned data optimization strategy, we refined the dataset to 8,691 structures, achieving a redundancy reduction of up to 55.98%. For a closer look at the optimized dataset's traits, refer to **Figure S2**, which presents the distribution of optimized structures in the α-Fe/H binary system under t-SNE visualization. Leveraging the outcomes of the visualization, we conducted a targeted data screening. As a result, we established an optimized subset comprising 8,691 structures, indicating a reduction in data volume by ~56.46%. **Figure 4b** illustrates the comparison of the average training time per iteration between the original and optimized datasets. The average training time per iteration decreased from 1,677 to 742.7 seconds, resulting in **an impressive 55.8% reduction in time cost per iteration**. This optimization significantly reduced the expenses incurred during the training process.

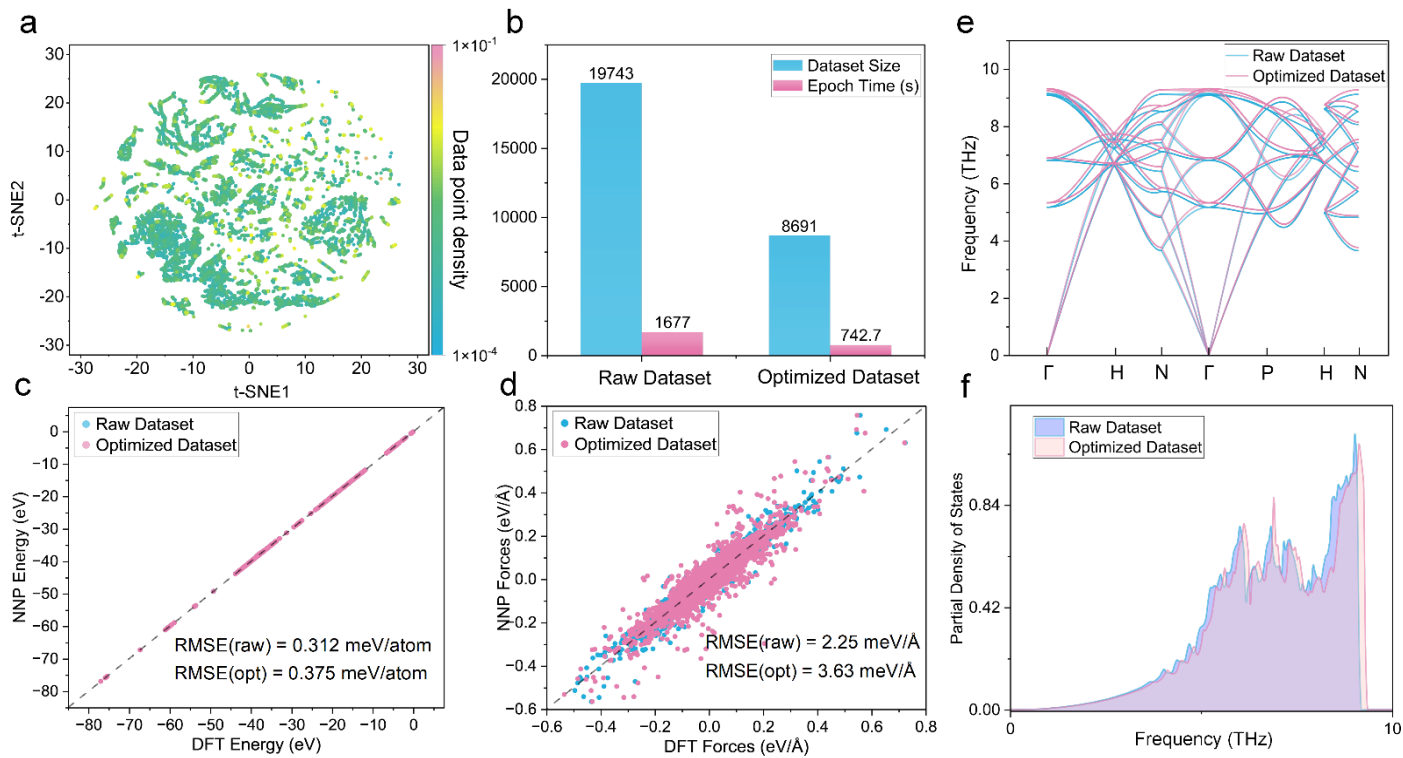

**Figure 4. Application and validation of the DV-LAE optimization method for the α-Fe/H binary system dataset, highlighting the significant reduction in training time and dataset size achieved while preserving prediction accuracy and key physical properties.** (a) The t-SNE visualization of α-Fe/H binary system dataset, highlighting data point density distribution after dimensionality reduction. (b) Average training time per iteration for the original dataset (19,743 structures) *versus* the optimized dataset (8,691 structures). (c-d) Comparison of the prediction errors in energies and forces: for the full dataset, the energy RMSE is 0.312 meV/atom and the force RMSE is 2.25 meV/Å. For the optimized dataset, the energy RMSE increases to 0.375 meV/atom, and the force RMSE rises to 3.63 meV/Å. (e-f) Phonon spectrum and band diagram comparison: the phonon spectrum and band diagram generated using the MLP trained on the full dataset show strong consistency with that produced from the optimized dataset, indicating the preservation of key physical insights.

To evaluate the effect of this dataset optimization on predictive capability, the performance of the MLP on the test set is depicted in **Figures 4c-d**. With the full dataset, the RMSE for energy prediction is 0.312 meV/atom, and 2.25 meV/Å for force prediction. However, when the model is trained on the optimized dataset, the RMSE for energy prediction increases to 0.375 meV/atom, and the force prediction rises to 3.63 meV/Å.

**Figures 4e-f** show the phonon spectrum and band structures generated using the MLP trained on both the raw dataset and the optimized dataset. The trends exhibited in both figures reveal a high degree of consistency, indicating that the optimization process has not significantly altered the underlying

physical properties represented by the phonons. This suggests that the optimization method employed is effective at preserving essential features while improving computational efficiency and accuracy. The phonon spectra in both figures reflect similar peaks and overall patterns, demonstrating that the critical vibrational characteristics of the material remain intact despite the reduction in dataset size. This consistency reinforces the notion that the MLP maintains its reliability and robustness across varying datasets.

**Table 1.** Comparison among the MLPs' performance for the Fe-H binary systems.

| Properties | | OPT | DFT | REF | ALL |
|---|---|---|---|---|---|
| Lattice constants (Å) | $a_0$ | 2.83 | 2.83 | 2.83 | 2.83 |
| Vacancy formation energy (eV) | monovacancy | 2.31 | 2.22 | 2.20 | 2.30 |
| Elastic constants (GPa) | $C_{11}$, | 290.2 | 296.0 | 296.0 | 270.1 |
| Elastic constants (GPa) | $C_{12}$ | 147.8 | 151.0 | 147.0 | 120.3 |
| Elastic constants (GPa) | $C_{14}$ | 92.8 | 105.0 | 96.0 | 90.2 |

Note: OPT: MLP trained with optimized dataset (this work); DFT: density functional theory calculation results; REF: literature MLP results from Ref.36; ALL: MLP trained with full unoptimized dataset.

**Table 1** presents a comparative analysis of the MLP performance for the Fe-H binary systems. We compared the properties obtained from the MLP trained with the optimized dataset (OPT) against those trained with the raw dataset, calculated *via* DFT, and obtained from the reference literature (REF)[36]. Specifically, OPT represents the properties calculated from the MLP trained with the optimized dataset using the method presented in this paper, DFT denotes the properties derived from first-principles calculations, and REF indicates the properties calculated from the MLP trained by the authors of the reference literature.[36] To avoid issues arising from training configurations, we also established a set labeled "ALL", which represents the properties calculated from the MLP function trained with the full dataset under the same configuration. The results show that the lattice constants provided by OPT, ALL, DFT, and REF are identical, at 2.83 Å, indicating that the results are consistent whether obtained through training with the optimized dataset, the full dataset, or *via* first-principles calculations. In terms of vacancy formation energy, there are slight differences among the values provided by OPT, ALL,

DFT, and REF, but all fall within the range of 2.20 to 2.31 eV, with overall small discrepancies. Regarding the elastic constants C11, C12, and C14, there are some differences between the results given by OPT and those by DFT and REF, but the results of OPT and ALL are relatively close to each other. This suggests that under the same training configurations, the optimized model demonstrates a high degree of consistency in describing the elastic behavior of the Fe-H binary systems.

2.4 Comparison Between the SOAP-based Similarity Screening and the DV-LAE Method for Data Screening and Prediction Accuracy

For comparison, a data screening strategy based on similarity assessment using the Smooth Overlap of Atomic Positions (SOAP) descriptor was evaluated.[37] Specifically, in this SOAP-based screening approach, the atomic descriptors of the candidate structures and those in the training set were calculated using SOAP descriptors. SOAP descriptors effectively capture the high-dimensional representations of atomic environments, providing a robust tool for similarity assessment. The steps of the similarity calculation within this approach are as follows:

i) For each candidate structure, we calculate the representation vector of its atomic environment using SOAP descriptors.
ii) Similarly, SOAP descriptors are calculated for every structure in the training set.
iii) The Euclidean distance is then used to compute the similarity between the descriptors of candidate structures and those in the training set. Specifically, for each descriptor in the training set, the minimum Euclidean distance to the candidate structure's descriptor is taken as the measure of similarity.
iv) Based on these similarity values, the maximum similarity and the average similarity are computed to assess how well the candidate structure represents the training set as a whole.

During the data screening process using this SOAP-based similarity evaluation, a similarity threshold was set. If the average similarity of a candidate structure exceeds this threshold, it is deemed sufficiently representative of the atomic environments in the training set and is included in the training data. This method allows for low-cost data screening without the need for expensive DFT calculations, as only structural information is required to compute SOAP descriptors and similarity scores. **However, a significant drawback of SOAP is that as the size of the training data increases, the computational cost of this similarity-based screening method also rises significantly.**

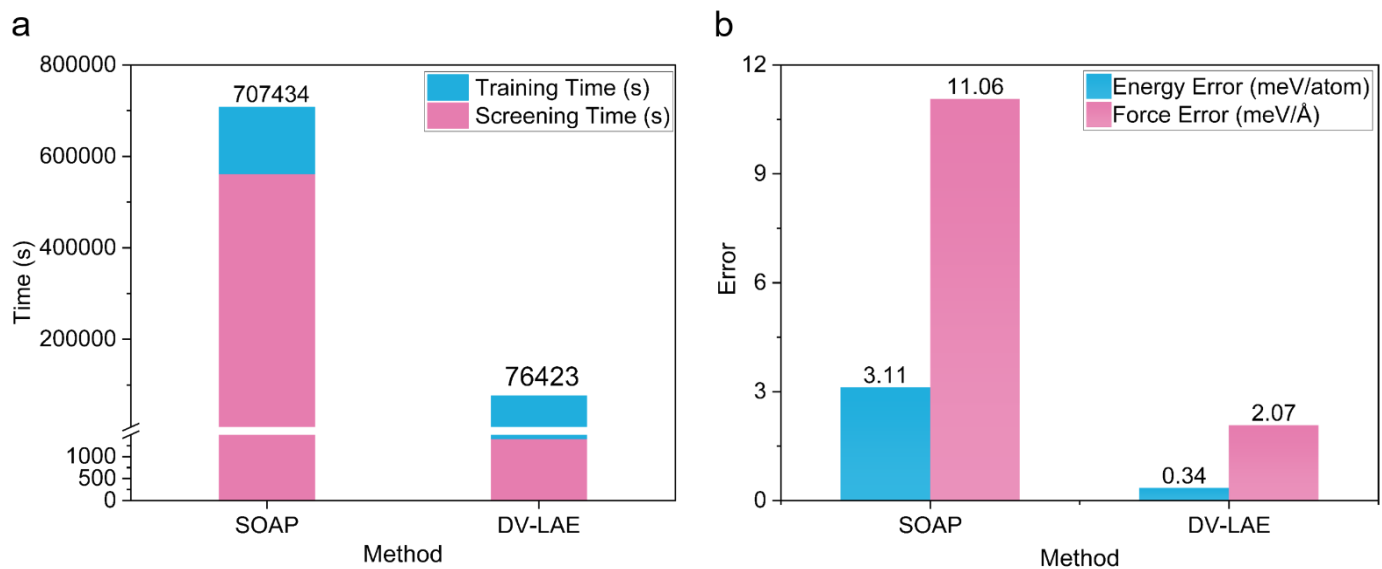

**Figure 5. Comparison between the SOAP-based screening method and the DV-LAE method on the Fe-H dataset.** (a) Total time comparison between the SOAP-based screening method and the DV-LAE method. Blue represents screening time, while pink indicates training time. (b) Error comparison showing both energy errors (in blue) and force errors (in pink) for the SOAP-based screening method and the DV-LAE method.

To compare our approach with the SOAP-based screening method and DV-LAE method, we applied both methods to the Fe-H dataset, which initially contained 19,743 structures. Using the SOAP-based screening method with a threshold of 0.1, data screening was completed in 562,185 seconds, resulting in an optimized dataset containing 17,900 structures. This dataset was then used to train a potential function over 100 iterations, with an average training time of 1,452 seconds per iteration. **Figure 5a** clearly shows the significant time difference between the two methods. The total time for the SOAP-based screening method is represented in blue and pink bars, with the blue portion indicating screening time and the pink portion representing training time. The SOAP-based screening method not only took longer for screening but also for training, with the total time approaching 707,434 seconds.

In contrast, using our method (DV-LAE), the data screening process on the same dataset resulted in an optimized subset of 8,691 structures, completed with significantly less time consumption. As shown in **Figure 5a**, DV-LAE exhibits a drastically reduced time for both screening and training, with a total of 75,423 seconds. The same training configuration of 100 iterations was employed, but the average time per iteration was reduced to 750 seconds, representing a considerable improvement in efficiency compared to the SOAP-based screening method. In terms of the prediction accuracy, the test set results showed that with the SOAP method, the energy error was 3.11 meV/atom, and the force error was 11.06 meV/Å. However, with our method, the energy error decreased to 0.339 meV/atom, and the force error dropped to 2.07 meV/Å. These improvements in prediction accuracy are shown in **Figure 5b**, where the **DV-LAE method significantly outperforms the SOAP-based screening method**, as indicated by the lower error for both energy and force predictions.

In summary, while the SOAP-based screening method effectively reduces the data size and improves training efficiency, **Figure 5** clearly demonstrates that our approach not only provides an even more efficient data screening process but also yields much better predictive accuracy in both energy and force predictions.

2.5 Evaluating the Visual Detection and Prediction Reliability of Difference Vector Space for Out-of-Distribution Data (ODD)

This study utilizes AIMD simulations and nonlinear downscaling techniques to elucidate the dynamic properties and structural diversity characteristics of the hydrogen and carbon systems at the atomic scale. The dynamic evolution analysis for the hydrogen system demonstrates that the gaseous hydrogen atoms in the $MgH_2$ system simulated at an elevated temperature of 700 K exhibit a characteristic localized motion pattern (**Figure 6a**). Structural snapshots from the AIMD trajectory are represented by blue points in a reduced-dimensional space, with the trajectory's evolution path dynamically tracked by connecting lines and the brown dashed curve. The dense distribution of trajectory points is highlighted by the locally enlarged regions, and the highly convergent distribution of trajectories indicates that the conformational rise and fall of the hydrogen atom system is effectively suppressed and a stable dynamic equilibrium state is maintained under specific thermodynamic conditions.

The structural diversity of carbon-based systems was systematically investigated through the t-SNE dimensionality reduction applied to a multidimensional dataset encompassing crystalline phases (*e.g.*, diamond and graphene), defect configurations, surface-modified architectures, and nanotube morphologies (**Figure 6b**). The visualization exhibits pronounced spatial segregation, where light-blue clusters, representing bulk structures, are broadly dispersed across the feature space, reflecting carbon's adaptability to diverse bonding environments and molecular geometries. The use of chromatically defined clusters corresponds to distinct atomic configurations, thus explicitly delineating the interplay between graphite interlayer interactions, defect-induced lattice distortions, and other microscale structural modulations. The distinct clustering of atomic configurations within the space further emphasizes the diversity in carbon's structural forms and provides valuable insights into the relationships between these various configurations (refer to **Table S1** for the classification of carbon structures in the GAP-20 dataset and the corresponding molecular structure schematics).

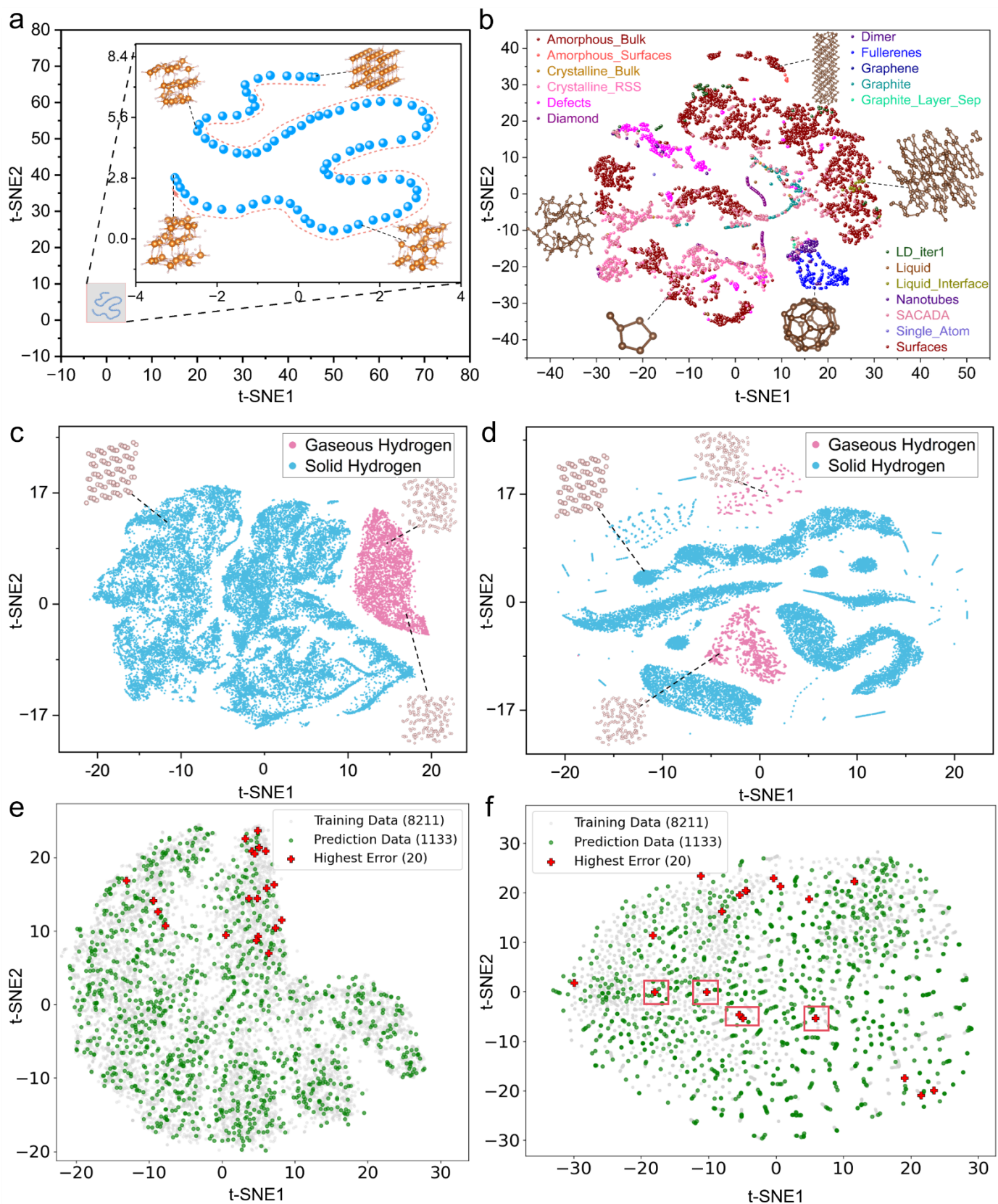

**Figure 6. Visualizations across various materials systems, demonstrating the representation of dynamic processes and diverse structural types, and the comparison of the effectiveness of DV-LAE and G-function representations for distinguishing phases.** (a) Visualization displays the AIMD trajectory of magnesium hydride systems. The hydrogen atoms within the magnesium hydrides are represented by grey spheres, the magnesium atoms are represented by brown spheres, and the dynamic movement of the molecules over time is depicted by the brown curve, which traces the

molecular trajectories. (b) Visualization of a dataset related to carbon structures. Different colors represent various configurations and forms of carbon atoms. (c-d) Comparison of hydrogen phase visualization: embeddings generated using (c) the DV-LAE representation and (d) a zero-padded G-function representation. (e-f) Comparison of feature space visualization for identifying potential OOD data: t-SNE embeddings generated using (e) DV-LAE features and (f) G-function features, displaying training data (light grey dots) and prediction data (green dots), along with the 20 highest error prediction points (red diamond). Red boxes in (f) highlight specific highest error points whose integration within regions of other data points in the G-function visualization makes their potential OOD status less apparent.

To further assess the capability of different feature representation methods in distinguishing structural phases and capturing subtle similarities, we analyzed a dataset containing 4,000 gaseous hydrogen and 25,439 solid hydrogen configurations. **Figure 6c** presents the two-dimensional embedding obtained using our DV-LAE method. It clearly separates the gaseous phase (pink points) from the solid phase (blue points). Crucially, the gaseous hydrogen configurations form a single, cohesive cluster, accurately reflecting the inherent structural similarity expected within this phase, despite variations in specific atomic positions. Representative atomic configurations linked from the clusters illustrate the disordered nature of the gas *versus* the more ordered solid structures. In contrast, **Figure 6d** shows the visualization results for the same dataset when the features are preprocessed using a standard technique involving G-function representation followed by zero-padding to handle variable system sizes, before applying dimensionality reduction. While this approach also successfully segregates the solid and gaseous phases, it problematically splits the gaseous hydrogen data into two distinct clusters. As indicated by the representative structures, these two separated gaseous clusters actually contain configurations that are physically highly similar. **This fragmentation suggests that the zero-padding preprocessing step, necessary for fixed-dimension input, may introduce artifacts or obscure the true similarity between these gaseous configurations in the feature space.** Visually, this manifests as the artificial splitting of the cohesive gaseous phase into separate clusters in **Figure 6d**, contrasting with the single, cohesive cluster in **Figure 6c**.

Therefore, comparing **Figures 6c** and **6d** highlights a key advantage of the DV-LAE method. By effectively handling variable structural environments without artificial padding, DV-LAE generates a feature representation that leads to a more physically meaningful clustering in the reduced-dimensional space. Its ability to group all structurally similar gaseous hydrogen configurations into a single cluster demonstrates superior fidelity in capturing the essential characteristics of the atomic environments compared to the baseline G-function padding approach, which leads to an artificial separation of similar structures. **This underscores the effectiveness of DV-LAE for accurate structural analysis and phase identification.**

Furthermore, to investigate the utility of feature space visualization in identifying potential out-of-distribution (OOD) data, we examine the spatial distribution of prediction points with the highest errors. This strategy can identify new structures that are poorly predicted by the model, indicating their OOD status because the model may not have encountered similar structures during training.

**Figures 6e** and **6f** provide vital spatial context by visualizing high-dimensional feature vectors in a two-dimensional t-SNE space, allowing for a comparison between our proposed DV-LAE method (**Figure 6e**) and a standard G-function representation with zero-padding (**Figure 6f**). Both panels display the training data (light grey dots) and prediction data (green dots), alongside the 20 prediction data points exhibiting the highest error (red diamond). A key observation when comparing the two is evident in **Figure 6e**: the highest error points identified using the DV-LAE feature space demonstrate a more pronounced spatial segregation. These points predominantly cluster in the upper-right and lower-right peripheral regions, appearing distinctly separated from the core of the training data distributions. In contrast, while the G-function representation in **Figure 6f** also shows these highest error points, their distribution can be misleading. For instance, some high-error points (demarcated by red boxes in **Figure 6f**), despite their high error suggesting potential OOD status, appear relatively integrated within the broader cloud of prediction data points or are even embedded within regions that seem to contain many other data points, rather than being clearly segregated at the periphery. This makes their potential OOD nature less apparent in the G-function based visualization (**Figure 6f**) compared to the clearer separation offered by DV-LAE (**Figure 6e**).

This direct comparison between **Figure 6e** (DV-LAE) and **Figure 6f** (G-function) underscores an advantage of the DV-LAE feature representation, as visualized in **Figure 6e**. The feature space learned by DV-LAE maps potentially problematic structures (those with high prediction error) to regions that are more clearly visually distinguishable in the low-dimensional embedding. This enhanced separation, evident in **Figure 6e**, provides a stronger qualitative indicator that these high-error points may indeed represent OOD samples. Notably, the visual appearance of **Figure 6f** (G-function) shows points that seem more overlapping or tightly clustered, and its highest error points (even those in red boxes) are less clearly segregated, compared to **Figure 6e** (DV-LAE). This observation is consistent with the G-function representation potentially mapping different samples to similar locations in the t-SNE space, resulting in a less discriminative visualization where distinct data subsets, including potential OOD samples, are less clearly separated. This facilitates a more effective visual assessment of potential OOD status with DV-LAE (**Figure 6e**) than with the G-function based approach (**Figure 6f**).

## Summary

In summary, we have introduced the novel DV-LAE method for analyzing atomic configurations and data redundancy in MLP training. Through targeted data screening, we successfully reduced data volume and training times while maintaining the high prediction accuracies of MLPs.

For the high-temperature, high-pressure hydrogen systems, an optimized dataset was created, resulting in a 31.79% reduction in data volume and a 29.0% decrease in average training time per iteration. The optimized dataset maintained comparable accuracy to the original, with only slight increases in the RMSE for energy and force predictions.

The α-Fe/H binary system dataset was also optimized, reducing data volume by 55.98% and training time by 55.8%. Despite increased RMSE values for energy and force predictions, the optimized dataset preserved key physical insights, as evidenced by consistent phonon spectra and band structures.

A comparison between the SOAP and DV-LAE methods for data screening revealed that the DV-LAE method not only reduced screening and training times significantly but also improved prediction accuracy. The DV-LAE method also provided a visualization tool to assess the rationality of newly simulated structures, identifying OOD data points that may compromise model performance.

Overall, the study demonstrated the effectiveness of data optimization techniques in enhancing computational efficiency and accuracy in materials science and chemistry applications, while also providing valuable insights into the relationships among various atomic configurations. The DV-LAE method emerged as a powerful tool for data screening and model evaluation, offering researchers a means to dynamically assess the trustworthiness of predictions for new, unseen data points. **Step-by-step guidance on implementing the DV-LAE methodology is available in Demonstration Videos within the associated GitHub repository.**

## Methods

First, a reference structure encompassing all elemental species in the target system was established as the computational benchmark. This representative configuration ensures that the reference structure can fully represent all elemental combinations and interactions within the target system. Adopting this approach provides a standardized environment for analyzing atomic interactions among different elements. Furthermore, the feature vectors derived from this reference structure are capable of accurately reflecting variations in the local environment of the system, thereby establishing a comprehensive basis for comparing and analyzing other structures.

Subsequently, atomic environment descriptors were subsequently employed to characterize local coordination features. These descriptors include radial symmetry functions (Equation 1) and angular symmetry functions (Equations 2-3), which provide a nuanced description of the system's local environment.[40,41] A cutoff function (Equation 4) is additionally applied to control the spatial range of the local environment captured by the aforementioned symmetry functions, with the coverage radius being determined by the specified cutoff parameter.

$$G_i^2 = \sum_j e^{-\eta(R_{ij}-R_s)^2} \cdot f_c(R_{ij}) \tag{1}$$

$$G_i^4 = 2^{1-\xi} \sum_{j,k \neq i}^{all} (1+\lambda\cos\theta_{ijk})^\xi \cdot e^{-\eta(R_{ij}^2+R_{ik}^2+R_{jk}^2)} \cdot f_c(R_{ij}) \cdot f_c(R_{ik}) \cdot f_c(R_{jk}) \tag{2}$$

$$G_i^5 = 2^{1-\xi} \sum_{j,k \neq i}^{all} (1+\lambda\cos\theta_{ijk})^\xi \cdot e^{-\eta(R_{ij}^2+R_{ik}^2)} \cdot f_c(R_{ij}) \cdot f_c(R_{ik}) \tag{3}$$

$$f_c(r) = \begin{cases} 1, & \text{for } 0 \leq r < r_{ci} \\ f(x), & \text{for } r_{ci} \leq r < r_c \text{ where } x = \dfrac{r-r_{ci}}{r_c - r_{ci}} \\ 0 & \text{for } r \geq r_c \end{cases} \tag{4}$$

MLP functions rely on these descriptors to predict the contribution of each atomic pair to the total energy of the system, thereby summing up to obtain the total energy of the system. Therefore, these descriptors have a significant impact on the accuracy of the MLP functions. By setting different parameter configurations, we can obtain various combinations of descriptors that reflect the diversity of the local environment of the system. However, since a single descriptor only describes the local environment, to differentiate between different systems, we need to consider the values of all descriptors for each structure comprehensively. To effectively capture the differences between structures, we employ histogram statistics to tally occurrences within each descriptor value interval. Inspired by the "less is more" concept highlighted in literature,[42] our method emphasizes the presence of values across intervals rather than their exact quantities, allowing for the efficient representation of structural features even with smaller training sets.

In the construction of difference vectors, we leverage a binary representation to identify disparities between the histogram intervals of descriptors in the current structure and those in a reference structure. This comparison is encapsulated by a binary function that flags differences, utilizing the XOR operation to yield a binary outcome where '1' signifies a difference and '0' indicates similarity.

The generation of the difference vector $D$ is conducted through a systematic process that leverages histogram statistics and binary comparison to encapsulate the disparities between the current structure and a reference structure across various descriptors $G_i$. The procedure is as follows:

i) **Histogram Construction:** For each descriptor $G_i$ of the current and reference structures, construct histograms by binning the values into a predefined number of intervals N (The selection of N is described in **Figures S6-S7**). Let $H_{current,j}$ and $H_{ref,j}$ represent the histogram bins for the current and reference structures, respectively, where $j$ denotes the interval index.

ii) **Binary Comparison *via* XOR Operation:** For each interval $j$, apply the XOR operation to the corresponding histogram bin counts to identify differences:

$$d_{i,j} = H_{current,j} \oplus H_{ref,j} \tag{5}$$

Here, $d_{i,j}$ is a binary variable where '0' indicates no difference (both bins have the same count) and '1' signifies a difference (bins have distinct counts).

iii) **Aggregation of Descriptor Differences:** For each descriptor $G_i$, aggregate the binary differences $d_{i,j}$ across all intervals to form a descriptor-specific difference vector $\vec{d_i}$:

$$\vec{d_i} = [d_{i,1}, d_{i,2}, \cdots, d_{i,N}] \tag{6}$$

iv) **Formation of the Comprehensive Difference Vector:** Concatenate the descriptor-specific difference vectors $\vec{d_i}$ for all descriptors $G_i$ in the system to create the comprehensive difference vector $D$:

$$D = [\vec{d_1}, \vec{d_1}, \cdots, \vec{d_m}] \tag{7}$$

where $m$ is the total number of descriptors.

This difference vector $D$ serves as a compact representation that captures the relative differences between structures. By applying dimensionality reduction techniques such as the t-SNE, we can visually represent these differences, facilitating a better understanding of structural variations.[30,31]

Ultimately, by strategically selecting a subset of data that reflects this difference vector analysis, we can curate a dataset that minimizes redundancy while preserving diversity. This refined dataset serves as a high-quality input for training MLPs, enhancing the efficiency and accuracy of our models without compromising on the breadth of structural representation.

**Code Availability**

Our developed codes in this study, and all the structures analyzed in this work (including high-temperature, high-pressure hydrogen systems, Fe–H binary (α-Fe/H) systems, magnesium hydride ($MgH_2$) systems, and carbon allotrope structures) are available *via*: https://github.com/Weijie-Yang/DV-LAE. **The repository also includes Tutorial Videos demonstrating the software usage and features.**

**Acknowledgements**

This work was funded by the Natural Science Foundation of Hebei (Grant No. E2020502023) and JSPS KAKENHI (Nos. JP25H01508, JP24K23068, and JP25K17991). We acknowledge the Center for Computational Materials Science, Institute for Materials Research, Tohoku University for the use of MASAMUNE-IMR (Nos. 202412-SCKXX-0211 and 202412-SCKXX-0209), and the Institute for Solid State Physics (ISSP) at the University of Tokyo for the use of their supercomputers.

**Conflict of Interests**

There is no conflict of interests to declare.

**Reference**

1. Sholl, D. S. & Steckel, J. A. *Density Functional Theory: A Practical Introduction*. (John Wiley & Sons, 2022).

2. Yang Yang *et al.* Taking materials dynamics to new extremes using machine learning interatomic potentials. *Journal of Materials Informatics* **1**, 10 (2021).

3. Liu, Z.-Q. *et al.* $FT^2DP$: large atomic model fine-tuned machine learning potential for accelerating atomistic simulation of iron-based Fischer-Tropsch synthesis. *Journal of Materials Informatics* **5**, (2025).

4. Wu, X., Kang, F., Duan, W. & Li, J. Density functional theory calculations: A powerful tool to simulate and design high-performance energy storage and conversion materials. *Progress in Natural Science: Materials International* **29**, 247–255 (2019).

5. Burke, K. Perspective on density functional theory. *The Journal of chemical physics* **136**, (2012).

6. Rupp, M., Ramakrishnan, R. & Von Lilienfeld, O. A. Machine learning for quantum mechanical properties of atoms in molecules. *The Journal of Physical Chemistry Letters* **6**, 3309–3313 (2015).

7. Ramprasad, R., Batra, R., Pilania, G., Mannodi-Kanakkithodi, A. & Kim, C. Machine learning in materials informatics: recent applications and prospects. *npj Computational Materials* **3**, 54 (2017).

8. Poltavsky, I. & Tkatchenko, A. Machine learning force fields: Recent advances and remaining challenges. *The journal of physical chemistry letters* **12**, 6551–6564 (2021).

9. Unke, O. T. *et al.* Machine learning force fields. *Chemical Reviews* **121**, 10142–10186 (2021).

10. Gubernatis, J. E. & Lookman, T. Machine learning in materials design and discovery: Examples from the present and suggestions for the future. *Physical Review Materials* **2**, 120301 (2018).

11. Mailoa, J. P. *et al.* A fast neural network approach for direct covariant forces prediction in complex multi-element extended systems. *Nat Mach Intell* **1**, 471–479 (2019).

12. Behler, J. Perspective: Machine learning potentials for atomistic simulations. *The Journal of Chemical Physics* **145**, 170901 (2016).

13. Bartók, A. P., Kermode, J., Bernstein, N. & Csányi, G. Machine learning a general-purpose interatomic potential for silicon. *Physical Review X* **8**, 041048 (2018).

14. Deringer, V. L. & Csányi, G. Machine learning based interatomic potential for amorphous carbon. *Physical Review B* **95**, 094203 (2017).

15. Behler, J. & Parrinello, M. Generalized Neural-Network Representation of High-Dimensional Potential-Energy Surfaces. *Phys. Rev. Lett.* **98**, 146401 (2007).

16. Behler, J. & Parrinello, M. Generalized neural-network representation of high-dimensional potential-energy surfaces. *Physical review letters* **98**, 146401 (2007).

17. Zhang, L., Han, J., Wang, H., Car, R. & E, W. Deep potential molecular dynamics: a scalable model with the accuracy of quantum mechanics. *Physical review letters* **120**, 143001 (2018).

18. Han, J., Zhang, L., Car, R., & others. Deep potential: A general representation of a many-body potential energy surface. *arXiv preprint arXiv:1707.01478* (2017).

19. Batatia, I., Kovacs, D. P., Simm, G., Ortner, C. & Csányi, G. MACE: Higher order equivariant message passing neural networks for fast and accurate force fields. *Advances in neural information processing systems* **35**, 11423–11436 (2022).

20. Musaelian, A. *et al.* Learning local equivariant representations for large-scale atomistic dynamics. *Nature Communications* **14**, 579 (2023).

21. Zhou, L., Pan, S., Wang, J. & Vasilakos, A. V. Machine learning on big data: Opportunities and challenges. *Neurocomputing* **237**, 350–361 (2017).

22. Jordan, M. I. & Mitchell, T. M. Machine learning: Trends, perspectives, and prospects. *Science* **349**, 255–260 (2015).

23. Wei, J. *et al.* Machine learning in materials science. *InfoMat* **1**, 338–358 (2019).

24. Chen, J. The development and comparison of molecular dynamics simulation and Monte Carlo simulation. in *IOP Conference Series: Earth and Environmental Science* vol. **128** 012110 (IOP Publishing, 2018).

25. Rupakheti, C., Virshup, A., Yang, W. & Beratan, D. N. Strategy To Discover Diverse Optimal Molecules in the Small Molecule Universe. *J. Chem. Inf. Model.* **55**, 529–537 (2015).

26. Mu, S. *et al.* Uncovering electron scattering mechanisms in NiFeCoCrMn derived concentrated solid solution and high entropy alloys. *npj Computational Materials* **5**, 1 (2019).

27. Bowman, J. M. *et al.* The MD17 datasets from the perspective of datasets for gas-phase "small" molecule potentials. *J. Chem. Phys.* **156**, 240901 (2022).

28. Zhang, L., Lin, D.-Y., Wang, H., Car, R. & E, W. Active learning of uniformly accurate interatomic potentials for materials simulation. *Physical Review Materials* **3**, 023804 (2019).

29. Greenacre, M. *et al.* Principal component analysis. *Nature Reviews Methods Primers* **2**, 100 (2022).

30. Van der Maaten, L. & Hinton, G. Visualizing data using t-SNE. *Journal of machine learning research* **9**, (2008).

31. Wattenberg, M., Viégas, F. & Johnson, I. How to use t-SNE effectively. *Distill* **1**, e2 (2016).

32. Zhang, Y. *et al.* DP-GEN: A concurrent learning platform for the generation of reliable deep learning based potential energy models. *Comput Phys Commun* **253**, 107206 (2020).

33. Behler, J. Atom-centered symmetry functions for constructing high-dimensional neural network potentials. *The Journal of chemical physics* **134**, (2011).

34. Cheng, B., Mazzola, G., Pickard, C. J. & Ceriotti, M. Evidence for supercritical behaviour of high-pressure liquid hydrogen. *Nature* **585**, 217–220 (2020).

35. Meng, F.-S. *et al.* General-purpose neural network interatomic potential for the α-iron and hydrogen binary system: Toward atomic-scale understanding of hydrogen embrittlement. *Physical Review Materials* **5**, 113606 (2021).

36. Zhang, S., Meng, F., Fu, R. & Ogata, S. Highly efficient and transferable interatomic potentials for α-iron and α-iron/hydrogen binary systems using deep neural networks. *Computational Materials Science* **235**, 112843 (2024).

37. Rowe, P., Deringer, V. L., Gasparotto, P., Csányi, G. & Michaelides, A. An accurate and transferable machine learning potential for carbon. *The Journal of Chemical Physics* **153**, 034702 (2020).

38. Silverman, B. W. Density Estimation for Statistics and Data Analysis. (Routledge, 2018).

39. Omee, S. S., Fu, N., Dong, R., Hu, M. & Hu, J. Structure-based out-of-distribution (OOD) materials property prediction: a benchmark study. *npj Comput Mater* **10**, 144 (2024).

40. Behler, J. & Parrinello, M. Generalized Neural-Network Representation of High-Dimensional Potential-Energy Surfaces. *Phys. Rev. Lett.* **98**, 146401 (2007).

41. Behler, J. Atom-centered symmetry functions for constructing high-dimensional neural network potentials. *The Journal of Chemical Physics* **134**, 074106 (2011).

42. Smith, J. S., Nebgen, B., Lubbers, N., Isayev, O. & Roitberg, A. E. Less is more: Sampling chemical space with active learning. *J. Chem. Phys.* **148**, 241733 (2018).